\documentclass[sigconf]{acmart}

\usepackage{url}
\AtBeginDocument{%
  }

\acmPrice{15.00}
\acmISBN{978-1-4503-XXXX-X/18/06}

\copyrightyear{2023}
\acmYear{2023}
\setcopyright{acmlicensed}\acmConference[MM '23]{Proceedings of the 31st ACM International Conference on Multimedia}{October 29-November 3, 2023}{Ottawa, ON, Canada}
\acmBooktitle{Proceedings of the 31st ACM International Conference on Multimedia (MM '23), October 29-November 3, 2023, Ottawa, ON, Canada}
\acmPrice{15.00}
\acmDOI{10.1145/3581783.3612524}
\acmISBN{979-8-4007-0108-5/23/10}

\begin{document}

\title{ControlStyle: Text-Driven Stylized Image Generation Using Diffusion Priors}

\author{Jingwen Chen}
\affiliation{%
  \institution{Sun Yat-sen University}
  \city{Guangzhou}
  \country{China}
}
\email{chenjingwen.sysu@gmail.com}
\author{Yingwei Pan}
\affiliation{%
  \institution{University of Science and Technology of China}
  \city{Hefei}
  \country{China}
}
\email{panyw.ustc@gmail.com}

\author{Ting Yao}
\affiliation{%
  \institution{HiDream.ai Inc.}
  \city{Beijing}
  \country{China}
}
\email{tingyao.ustc@gmail.com}

\author{Tao Mei}
\affiliation{%
  \institution{HiDream.ai Inc.}
  \city{Beijing}
  \country{China}
}
\email{tmei@hidream.ai}


\renewcommand{\shortauthors}{Jingwen Chen, Yingwei Pan, Ting Yao, \& Tao Mei}
\begin{abstract}
Recently, the multimedia community has witnessed the rise of diffusion models trained on large-scale multi-modal data for visual content creation, particularly in the field of text-to-image generation. In this paper, we propose a new task for ``stylizing’’ text-to-image models, namely text-driven stylized image generation, that further enhances editability in content creation. Given input text prompt and style image, this task aims to produce stylized images which are both semantically relevant to input text prompt and meanwhile aligned with the style image in style. To achieve this, we present a new diffusion model (ControlStyle) via upgrading a pre-trained text-to-image model with a trainable modulation network enabling more conditions of text prompts and style images. Moreover, diffusion style and content regularizations are simultaneously introduced to facilitate the learning of this modulation network with these diffusion priors, pursuing high-quality stylized text-to-image generation. Extensive experiments demonstrate the effectiveness of our ControlStyle in producing more visually pleasing and artistic results, surpassing a simple combination of text-to-image model and conventional style transfer techniques.

\end{abstract}

\ccsdesc[500]{Computing methodologies~Computer vision tasks}

\keywords{diffusion models, text-to-image generation, style transfer}

\begin{teaserfigure}
  \centering
  \vspace{-0.15in}
  \includegraphics[width=\textwidth]{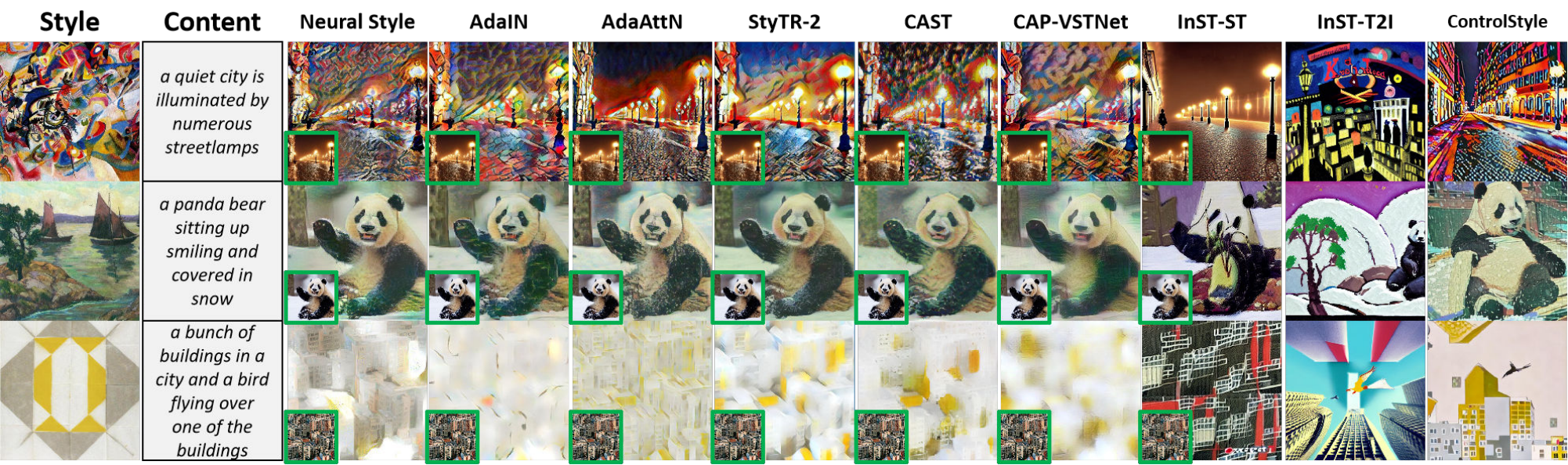}
  \vspace{-0.25in}
  \caption{In this work, we explore a new task of text-driven stylized image generation, i.e., directly generating stylized images based on style images and text prompts that describe the content. A simple solution for this task is to combine a text-to-image model (text $\Rightarrow$ image) and a style transfer network (content image $\Rightarrow$ stylized image) in a two-stage manner. In contrast, our ControlStyle unifies both stages into one diffusion process, leading to high-fidelity stylized images with better visual quality.}
  \label{fig:teaser}
\end{teaserfigure}

\maketitle

\section{Introduction}
Neural style transfer, a prominent research topic in multimedia and vision fields, aims to render an image with a desired style while preserving the underlying content. Pioneer researches \cite{perceptual,neural_style} achieve this goal by exploring the correlation between the features of content and style images extracted by a pre-trained convolutional neural networks. Follow-up works \cite{adain,dain} propose to transform the features of the content image to those ones that are aligned with the style image in global/local statistics (e.g., mean and variance) for arbitrary style transfer. Later, GAN-based methods \cite{cyclegan,cut} have been developed to tackle the barrier of training style transfer network with unpaired content and style images resting on adversarial learning.

While promising results are achieved, typical style transfer methods belong to image-to-image translation in single modality and require the inputs of content image and style image. To further eliminate the need of content image and enhance user editability, we develop a new task of text-driven stylized image generation. In this task, the system is required to generate a high-quality stylized image that is both semantically aligned with an input text prompt and consistent with an input style image in style. Recently, remarkable advancements in text-to-image synthesis \cite{imagen,glide,ldm:cvpr22} have been attained by diffusion models \cite{ddpm:ho20}. ControlNet \cite{controlnet} further incorporates additional conditions, such as edge images or depth maps, into a pre-trained text-to-image diffusion model to better control the spatial structure of the generated samples. Inspired by these works, we propose a new diffusion model, namely ControlStyle, to resolve the new task of text-driven stylized image generation.

Technically, our ControlStyle upgrades a pre-trained text-to-image diffusion model with a trainable modulation network, enabling more conditions of text prompts and style images for better editability. Specifically, the modulation network is initialized from the U-Net of a diffusion model along side with one more condition, i.e., style image, aiming to produce structurally and semantically aware style features. These style features are connected back to the U-Net through zero convolutional layers to modulate the behavior of U-Net for text-driven stylized image generation. During the training procedure, the pre-trained diffusion model is frozen to preserve the strong capability of text-to-image generation learnt from billions of multimodal image-text data, while the modulation network is optimized to stylize the pre-trained diffusion model.

Note that because there is no underlying correlation between the image-text pair and the style image, it is not trivial to train our ControlStyle under such unpaired setting with image-text pairs and another set of arbitrary style images. In an effort to mitigate this problem, we devise novel diffusion style and content regularizations to facilitate the optimization of ControlStyle. The diffusion style regularization enforces the generated image to exhibit style consistency with the input style image, while the diffusion content regularization prevents the spatial structure from being heavily destroyed in the presence of the diffusion style regularization. Extensive experiments demonstrate that our ControlStyle surpasses a simple combination of text-to-image model and conventional style transfer techniques (see the examples in Figure \ref{fig:teaser}).

To summarize, the main contributions of this work are as follows: 1) A new task of text-driven stylized image generation is introduced, which aims to improve the editability of content creation. 2) A new diffusion model ControlStyle is proposed to stylize a pre-trained text-to-image diffusion model via a trainable modulation network. 3) Two key ingredients of optimizing ControlStyle are devised: diffusion style and content regularizations, which comprises the recipe of training diffusion models under the unpaired setting.

\section{Related Work}
\subsection{Neural Style Transfer}
\label{sec:style_transfer}
Neural style transfer \cite{neural_style,perceptual} is an appealing research topic in computer vision, which aims to render a content image in the style of another image. One of the pioneer works iteratively optimizes image pixels by matching deep image representations derived from Convolutional Neural Networks between the content image and the style image for artistic style transfer \cite{neural_style}. This work is later extended \cite{gaty_control} to decompose the style into several essential factors for more flexible manipulations in spatial location, scale and color. However, these optimization-based methods suffer from slow inference. To facilitate style transfer in real-time applications, Johnson \emph{et al.} combine the benefits of effective optimization with perceptual loss and high inference efficiency of training a image transformation feed-forward network in \cite{perceptual}. Since the style transfer model is trained on a pre-defined set of style images, the generalizability to arbitrary styles unseen in the training set is limited. In response to this limitation, a multitude of normalization-based methodologies \cite{adain,dain,adaattn} have been proposed to match the global/local statistics (e.g., mean and variance) of the content image and the style image. Later, inspired by adversarial generative models \cite{pan2017create,chen2019mocycle}, an innovative family of \cite{gan,pix2pix,cyclegan,cut} employs a minimax two-player game to facilitate the training of style transfer network in an unpaired setting, which mitigates the demand of paired content and style images. Though the aforementioned works can produce high-quality stylized images, a content image and a style image are required from the users. Several recent studies \cite{kwon2021clipstyler,fu2022language,tistyler,zeroshot_tistyler} contend that in certain scenarios, acquiring a desired style image may be challenging. Therefore, the task of text-guided image style transfer has been introduced, which substitutes the style image with a natural language sentence that conveys the desired stylistic attributes. For example, CLIPStyler \cite{kwon2021clipstyler} leverages the pre-trained text-image embedding model of CLIP \cite{clip} to align the stylized image and the input style prompt in the learned multimodal embedding space. Furthermore, DiffusionCLIP \cite{kim2021diffusionclip} extends CLIPStyler by employing a pre-trained diffusion model as the image generator for high-quality image synthesis.

\subsection{Diffusion Models}
In recent times, diffusion denoising probabilistic models (DDPM) \cite{ddpm:ho20} have engendered a remarkable breakthrough in the evolution of computer vision, particularly in related fields of image synthesis. DDPM can be formulated as a diffusion process and a reverse process. In the diffusion process, the data is subjected to incremental perturbations by Gaussian noise, eventually turning into pure Gaussian noise after hundreds of thousands of  steps. Conversely, in the reverse process, DDPM learns to recover the data by predicting the added noise and removing it progressively. Despite the impressive results achieved by DDPM in generative modeling, some drawbacks have hindered its application, such as high demands of computation resources and low inference speed. Considerable works \cite{ddim,ldm:cvpr22,ho2022classifier,classifierguidance} have been proposed to improve DDPMs and further tap its potentials. Given that direct optimization of DDPM in pixel space is computationally expensive, an alternative approach, latent diffusion models (LDM) \cite{ldm:cvpr22}, performs the training of DDPM in latent space learned by an auto-encoder, which makes the training and inference of DDPM much more efficient and produces stunning images. Due to these notable advancements, DDPM has manifested itself as the emerging trend in the realms of text-to-image synthesis \cite{imagen,glide,instructpix2pix,p2p}, 3D generation \cite{dreamfusion,makeit3d}, video generation \cite{ho2022imagenvideo,ho2022video}. For example, stable diffusion \cite{ldm:cvpr22} amalgamates the strengths of both pre-trained text-image embedding model (CLIP) and latent diffusion model in high-fidelity and fast image generation. Most recently, a new architecture ControlNet \cite{controlnet} is proposed to control the pre-trained stable diffusion with more conditions for text-to-image synthesis.

\subsection{Summary}
In this paper, we consider a new task of text-driven stylized image generation. Compared to the aforementioned text-guided image style transfer, this new task unleashes the need for a content image and an accurate description of a desired style. In text-driven stylized image generation, only an input text prompt and a style image are required to generate images that are both semantically aligned with the input text prompt and consistent with the style image in style. To resolve this problem, we propose a new diffusion model ControlStyle by upgrading a pre-trained text-to-image model with a trainable modulation network that enables more conditions of style images. Moreover, both diffusion style and content regularizations are designed to facilitate the training of ControlStyle under an unpaired setting (image-text pairs plus another set of style images).

\section{Approach}
\begin{figure*}[th]
  \centering
  \includegraphics[width=0.95\linewidth]{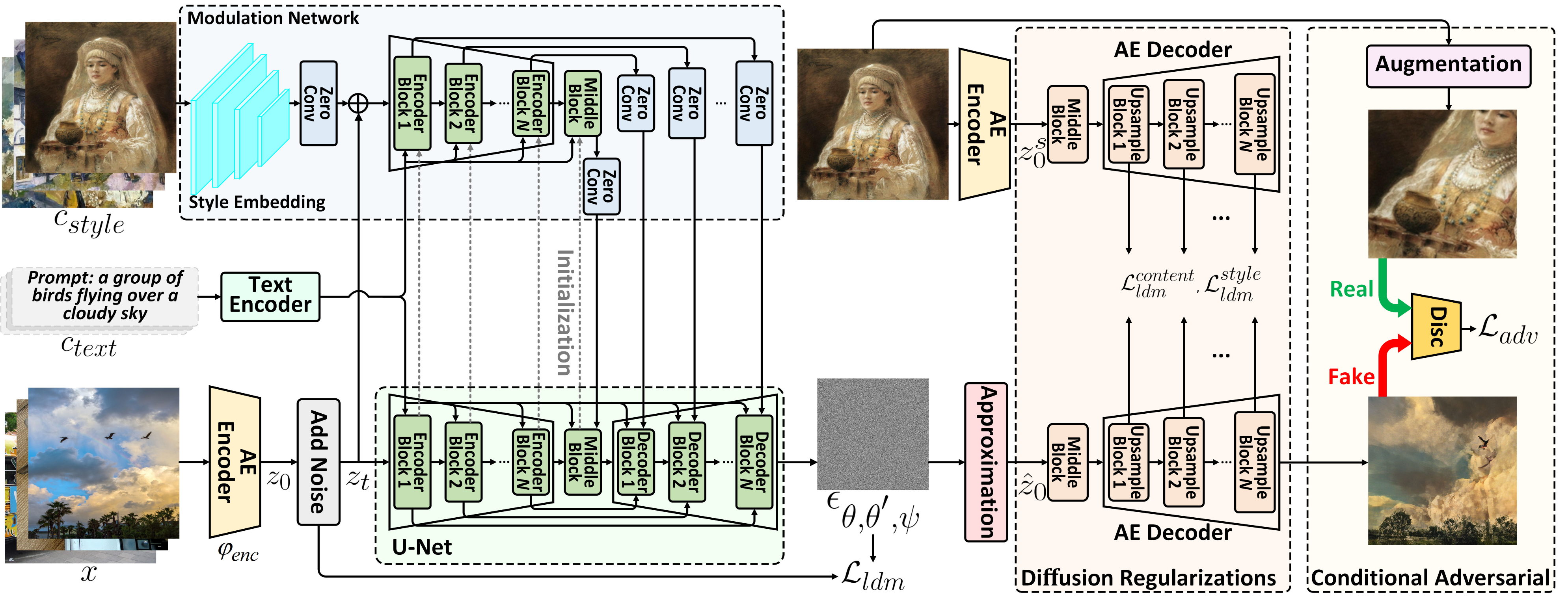}
  \vspace{-0.1in}
  \caption{Overall framework of ControlStyle. In ControlStyle, a trainable modulation network is initialized from U-Net of the frozen stable diffusion model and connected back to it through zero convolutions, which enables more conditions of style images. Specifically, content image $x$, prompt $c_{text}$ and style image $c_{style}$ are first encoded into low-dimension embeddings, respectively. Then, U-Net takes the noised latent embedding $z_t$ of $x$ and $c_{text}$ as inputs to produce multimodal features. Meanwhile, these inputs along with the extra condition $c_{style}$ are consumed by the modulation network to generate style features, which is further incorporated into U-Net for noise prediction via zero convolutions. Besides training ControlStyle with conventional diffusion loss $\mathcal{L}_{ldm}$, diffusion regularizations ($\mathcal{L}_{ldm}^{style}$ and $\mathcal{L}_{ldm}^{content}$) are proposed to effectively leverage image priors in the auto-encoder of stable diffusion. Moreover, a conditional adversarial loss $\mathcal{L}_{adv}$ is exploited to further boost visual quality. }
  \label{fig:framework}
  \vspace{-0.1in}
\end{figure*}

A vanilla solution to text-driven stylized image generation is to simply cascade a pre-trained text-to-image diffusion model (text $\Rightarrow$ content image) with a conventional style transfer technique (content image + style image $\Rightarrow$ stylized image). Nevertheless, this two-stage method underuses the image priors inherent in the diffusion model for content creation, and meanwhile ignore the interaction between content image generation and the stylization process. To mitigate these issues, we present a new framework, namely ControlStyle, which is an upgraded diffusion model with a trainable modulation network that jointly enables multiple conditions of text prompts and style images. In this section, we first briefly introduce the background of latent diffusion model in Section \ref{sec:tech_background}. Later, taking the publicly available text-to-image diffusion model (stable diffusion) as an example, the technical details of our ControlStyle are elaborated in Section \ref{sec:controlstyle} and \ref{sec:diffusion_reg}. Finally, the general training objective is demonstrated in Section \ref{sec:tech_training}.

\subsection{Background}
\label{sec:tech_background}
Diffusion probabilistic model (DDPM) \cite{ddpm:ho20} can be classified as a type of generative model, which is a parameterized Markov chain optimized to produce samples matching a target data distribution within finite timesteps $T$. In general, DDPM gradually adds noise to the data and finally destroys the data in compliance with a predefined variance schedule $\{\beta_t\}_1^T$ in a forward diffusion process. Conversely, in the reverse process, DDPM endeavors to reconstruct the original data by predicting the added noise and remove it in a progressive manner. Specifically, given the input data $x$ (also denoted as $x_0$), the noisy sample $x_t$ at an arbitrary timestep $t$ can be derived by 
\begin{equation}
\label{eq:ddpm_forward}
x_t = \sqrt{\bar{\alpha}_t} x_0 + \sqrt{1 - \bar{\alpha}_t} \epsilon,
\end{equation}
where $\alpha_t = 1 - \beta_t$, $\bar{\alpha}_t = \prod \limits_{s=1}^t \alpha_s$ and $\epsilon \sim \mathcal{N}({\bf{0,I}})$. Then, the sample $x_{t-1}$ can be recovered from $x_t$ by removing the predicted noise from DDPM (parameterized by $\epsilon_\theta$):
\begin{equation}
x_{t-1} = \frac{1}{\sqrt{\alpha_t}} (x_t - \frac{1 - \alpha_t}{\sqrt{1 - \bar{\alpha}_t}} \epsilon_\theta(x_t, t, c)) + \sigma_t \epsilon,
\end{equation}
where $c$ is some kind of condition (e.g., text for text-to-image generation via attention mechanism widely adopted in Vision Transformers \cite{yao2023dual,li2022contextual,yao2022wave}). Starting from a random noise $x_T \sim \mathcal{N}({\bf{0,I}})$, we can progressively execute the operation over the full chain with $T$ timesteps to produce a sample. Finally, the training objective for $\epsilon_\theta$ can be simply formulated as:
\begin{equation}
\label{eq:ddpm}
\mathcal{L}_{ddpm}(\theta, x) = \mathbb{E}_{t\sim \mathcal{U}(0,1), \epsilon \sim \mathcal{N}(0, I)} [w(t)\parallel\epsilon_\theta(x_t,t,c) - \epsilon \parallel_2^2],
\end{equation}
where $w(t)$ is a weighting function that depends on the timestep $t$.

Despite the capacity of the diffusion probabilistic model (DDPM) to achieve stable training and high-quality image generation \cite{ddpm:ho20,imagen,eDiff}, optimizing such models in pixel space often requires lots of GPU resources and the inference is also computationally expensive. To resolve these problems, Latent Diffusion Model (LDM) \cite{ldm:cvpr22} is proposed to learn DDPM in latent space, and achieves comparable even better results. To be more specific, in contrast to the previously discussed DDPM, LDM introduces an additional auto-encoder that has been pre-trained to project the image from pixel space $x \sim \mathcal{X}$ to latent space $z \sim \mathcal{Z}$ with an encoder $\varphi_{enc}$ and recover it to $x$ with a decoder $\varphi_{dec}$. In this way, $\epsilon_\theta$ can be trained in the latent space $\mathcal{Z}$, which has much smaller dimentionality than $\mathcal{X}$. Thus, the training objective Eq. (\ref{eq:ddpm}) can be rewritten as:
\begin{equation}
\label{eq:ldm_loss}
\begin{aligned}
\mathcal{L}_{ldm}(\theta, x) =~\mathbb{E}_{t\sim \mathcal{U}(0,T), \epsilon \sim \mathcal{N}(0, I)} [w(t)\parallel\epsilon_\theta(z_t,t,c) - \epsilon \parallel_2^2],
\end{aligned}
\end{equation}
where $z_t = \sqrt{\bar{\alpha}_t} \varphi_{enc}(x)+ \sqrt{1 - \bar{\alpha}_t} \epsilon$ .

\subsection{ControlStyle}
\label{sec:controlstyle}


In pursuit of high-quality text-driven stylized image generation, we design ControlStyle that unifies both text-to-image generation and image stylization into an end-to-end framework. Note that herein we use the publicly released stable diffusion as the pre-trained text-to-image diffusion model for training efficiency and reproducibility. In brief, stable diffusion comprises an auto-encoder, a text encoder and a U-Net \cite{unet}, for image encoding/decoding ($512 \times 512 \Leftrightarrow 64 \times 64$ ), text encoding and noise prediction, respectively. Inspired by ControlNet \cite{controlnet}, ControlStyle is designed to stylize the pre-trained stable diffusion model with a trainable modulation network. The modulation network consumes the input text $c_{text}$, the style image $c_{style}$ as well as the noisy latent code $z_t$ to produce style features that are both structurally and semantically aware of the inputs. These style features are utilized to modulate the pre-trained stable diffusion model for text-driven stylized image generation. In the learning process, only the modulation network is trained while the parameters of the pre-trained stable diffusion model are not tuned in order to preserve the strong text-to-image capability learned from billions of image-text data.

Specifically, the trainable modulation network is initialized from the encoder and middle blocks of U-Net in stable diffusion, and connected to the decoder blocks of U-Net through zero convolutional layers. It is worth mentioning that zero convolutional layer is a special convolutional layer with weights and bias initialized to zeros. Throughout the training process, parameters of these layers gradually transition from zeros to optimized values to avoid overfitting. Take a simple pre-trained model $f(\cdot)$ with only one neural network block as an example, the output can be denoted as
\begin{equation}
b = f(a).
\end{equation} 
After coupling $f(\cdot)$ with a trainable modulation block enabling an additional condition $c$, the new output can be derived as 
\begin{equation}
\widetilde{b} = f(a) + \psi^1(f^{'}(\psi^0(c))),
\end{equation}
where $\psi^0$ and $\psi^1$ are two zero convolutional layers, and $f^{'}(\cdot)$ is the trainable copy from $f(\cdot)$. The modulation network in ControlStyle is also formulated in a similar way. Let $\epsilon_\theta^{enc_{(i)}}$ be the $i$-th encoder block in U-Net and $\epsilon_\theta^{dec_{(j)}}$ be the \emph{symmetric} decoder block in U-Net, the output of the decoder block is originally computed as:
\begin{equation}
\small
\epsilon_\theta^{dec_{(j+1)}}(z_t,t,c_{text}) = \epsilon_\theta^{dec_{(j)}}(z_t,t,c_{text})+ \epsilon_\theta^{enc_{(i)}}(z_t,t,c_{text}).
\end{equation}
In ControlStyle, the output is modulated with one more condition of style image ($c_{style}$) as:
\begin{equation}
\small
\begin{aligned}
\epsilon_{\theta,\theta^{'},\psi}^{dec_{(j+1)}}(z_t,t,c_{text},c_{style}) =\ &\epsilon_\theta^{dec_{(j)}}(z_t,t,c_{text}) + \epsilon_\theta^{enc_{(i)}}(z_t,t,c_{text}) \\
&+ \psi_j^1( \epsilon_{\theta^{'}}^{enc_{(i)}}(z_t,t,c_{text},\psi^0(c_{style}))),
\end{aligned}
\end{equation}
where $\psi^0$ is the zero convolutional layer right before the modulation network, and $\psi_j^1$ is the zero convolutional layer for the $j$-th modulation block that connects the modulation block back to the decoder block of the frozen U-Net in stable diffusion. Additionally, to match the convolution size of U-Net, a style embedding network is devised to convert the stye image from $512 \times 512$ to $64 \times 64$. The overall framework of ControlStyle is illustrated in Figure. \ref{fig:framework}.

To ensure that the modulated output conforms to the distribution of the pre-trained stable diffusion, ControlStyle is trained with the conventional diffusion loss described in Eq. (\ref{eq:ldm_loss}) using a dataset of image-text pairs (e.g., MS-COCO).

\subsection{Diffusion Regularizations}
\label{sec:diffusion_reg}
For text-driven stylized image generation, the trained model is required to generate images that are both semantically aligned to the input text prompt and meanwhile consistent with the style image in style. To achieve these two goals, we design diffusion content and style regularizations to facilitate the learning of ControlStyle, which novelly leverages the image priors from the auto-encoder in the pre-trained stable diffusion model. Before performing the two diffusion regularizations, we reconstruct the clean sample $\hat{z}_0$ by \emph{approximation} following:
\begin{equation}
\hat{z}_0 = (z_t - \sqrt{1- \bar{\alpha}_t}\ \epsilon_{\theta,\theta^{'},\psi}(z_t,t,c_{text},c_{style})) / \sqrt{\bar{\alpha}_t}.
\end{equation}

\textbf{Diffusion style regularization.} The diffusion style regularization is designed to encourage the synthetic images share similar style as the input style image. In particular, we first feed $\hat{z}_0$ into the decoder of auto-encoder to produce intermediate features. For the target style, we first encode the input style image $c_{style}$ into a $64 \times 64$ latent code $z_0^s = \varphi_{enc}(c_{style})$ through the encoder of auto-encoder, and calculate the decoder features similarly. Next, the proposed diffusion style regularization aims to match the global statistics between intermediate features of $\hat{z}_0$ and $z_0^s$ for each upsample block. Let $\varphi_{dec}(\cdot)$ be the decoder in auto-encoder and $\varphi_{dec}^j(\cdot)$ be the $j$-th upsample block. Accordingly, the diffusion style regularization can be formulated as:
\begin{equation}
\small
\begin{aligned}
\mathcal{L}_{ldm}^{style}(\theta^{'},\psi,x,c_{text},c_{style}) = &\frac{1}{N} \sum_{j=1}^{N} (\parallel \mu(\varphi_{dec}^j(\hat{z}_0)) - \mu(\varphi_{dec}^j(z_0^s)) \parallel_2^2 \\
&+ \parallel \sigma(\varphi_{dec}^j(\hat{z}_0))^2 - \sigma(\varphi_{dec}^j(z_0^s))^2 \parallel_2^2),
\end{aligned}
\end{equation}
where $N$ is the number of upsample blocks, $\mu(\cdot)$ and $\sigma(\cdot)$ represent the mean and standard variance of inputs, respectively.

\textbf{Diffusion content regularization.} 
While the conventional diffusion loss encourages the semantic alignment between the generated image and the text prompt, the spatial structure may not be well preserved when solely using the diffusion style regularization during training. Therefore, another regularization (diffusion content regularization) is devised to prevent the structure being heavily destroyed by the style features from the modulation network. Similarly, we obtain the intermediate features from the decoder $\varphi_{dec}(\cdot)$ for $z_0$ (the latent code of the input content image $x_0$ during training) as described in diffusion style regularization. Then, the diffusion content regularization enforces the decoder features of $\hat{z}_0$ to spatially match with the ones of $z_0$, which can be defined as:
\begin{equation}
\mathcal{L}_{ldm}^{content}(\theta^{'},\psi,x,c_{text},c_{style}) =~ \frac{1}{CHW} \parallel \varphi_{dec}^J(\hat{z}_0) - \varphi_{dec}^J(z_0) \parallel_2^2,
\end{equation}
where $J$ indicates a specific upsample block (e.g., $UpBlock\_3$) in decoder, $CHW$ is the total number of elements in the feature map.

\subsection{Training}
\label{sec:tech_training}
Following the conventional training strategy in stable diffusion model, a dataset of image-text pairs (i.e., MS-COCO in this work) is required to optimize ControlStyle. Moreover, another set of style images (can be arbitrary styles) is also needed to modulate the pre-trained stable diffusion for text-driven stylized image generation. Besides the typical diffusion loss and the proposed diffusion regularizations, a conditional adversarial loss $\mathcal{L}_{adv}$ \cite{cgan} is also exploited to further improve the style learning: 
\begin{equation}
\begin{aligned}
\mathcal{L}_{adv}(\theta^{'},\psi,\xi,c_{text},c_{s})& =~ \mathbb{E}_{c_{s} \sim p_{data}(c_{s})}[log\ D_\xi(c_{s}^{aug}| c_{s})] \\
&+ \mathbb{E}_{\hat{x}_0 \sim \epsilon_{\theta,\theta^{'},\psi}}[log\ (1 - D_\xi(\hat{x}_0 | c_{s}))],
\end{aligned}
\end{equation}
where $c_s$ abbreviates $c_{style}$, $c_s^{aug}$ is an augmented sample from $c_{s}$, and $\hat{x}_0$ is the reconstructed image by feeding $\hat{z}_0$ to $\varphi_{dec}$. The final training objective is:
\begin{equation}
\mathcal{L}_{total} = \mathcal{L}_{ldm} + \mathcal{L}_{ldm}^{style} + \mathcal{L}_{ldm}^{content} + \mathcal{L}_{adv}.
\end{equation}
Once our ControlStyle is trained in such unpaired setting, we can generate an image of desired content and style conditioned on an input text prompt and a style image.

\section{Experiments}
In this section, we evaluate our ControlStyle in the new task of text-driven stylized image generation and compare it against generate-then-transfer methods \cite{neural_style,adain,adaattn,stytr2,cast,cap} and diffusion-based model \cite{inst}. Extensive experiments demonstrate the effectiveness of our ControlStyle in producing more visually pleasing results. Furthermore, we delve into our proposed diffusion regularizations and evaluate the effectiveness of our ControlStyle when generalized to unseen styles. Lastly, we upgrade ControlStyle by incorporating more controls (e.g., edge image) in a training-free manner to show its potentials in more interesting applications.

\subsection{Implementation Details}
Our ControlStyle is trained in an unpaired setting by using a common image-text dataset (MS-COCO \cite{mscoco}) and a large-scale painting dataset (WikiArt \cite{wikiart}). For image-text training pairs, we take about 118K and 5K from MS-COCO for training and validation, respectively. For arbitrary style images, about 60K style images from WikiArt are adopted for training and the remaining are utilized for validation. During training, each image-text pair is coupled with a randomly sampled style image to form a triplet $(x, c_{text}, c_{style})$. ControlStyle is optimized by Adam \cite{adam} with initial learning 0.0001 for about 60K iterations. Batch size is set to 4 and the input image resolution is set as $512 \times 512$. For diffusion style and content regularizations, features from the $UpBlock\_3$ and $UpBlock\_1,2,3$ in the auto-encoder of stable diffusion are exploited in our experiments. The adversarial discriminator model in $\mathcal{L}_{adv}$ is mainly implemented based on the codebase\footnote{https://github.com/NVIDIA/pix2pixHD/tree/master} \cite{wang2018pix2pixHD}.

\begin{figure*}[t]
  \centering
  \includegraphics[width=0.98\linewidth]{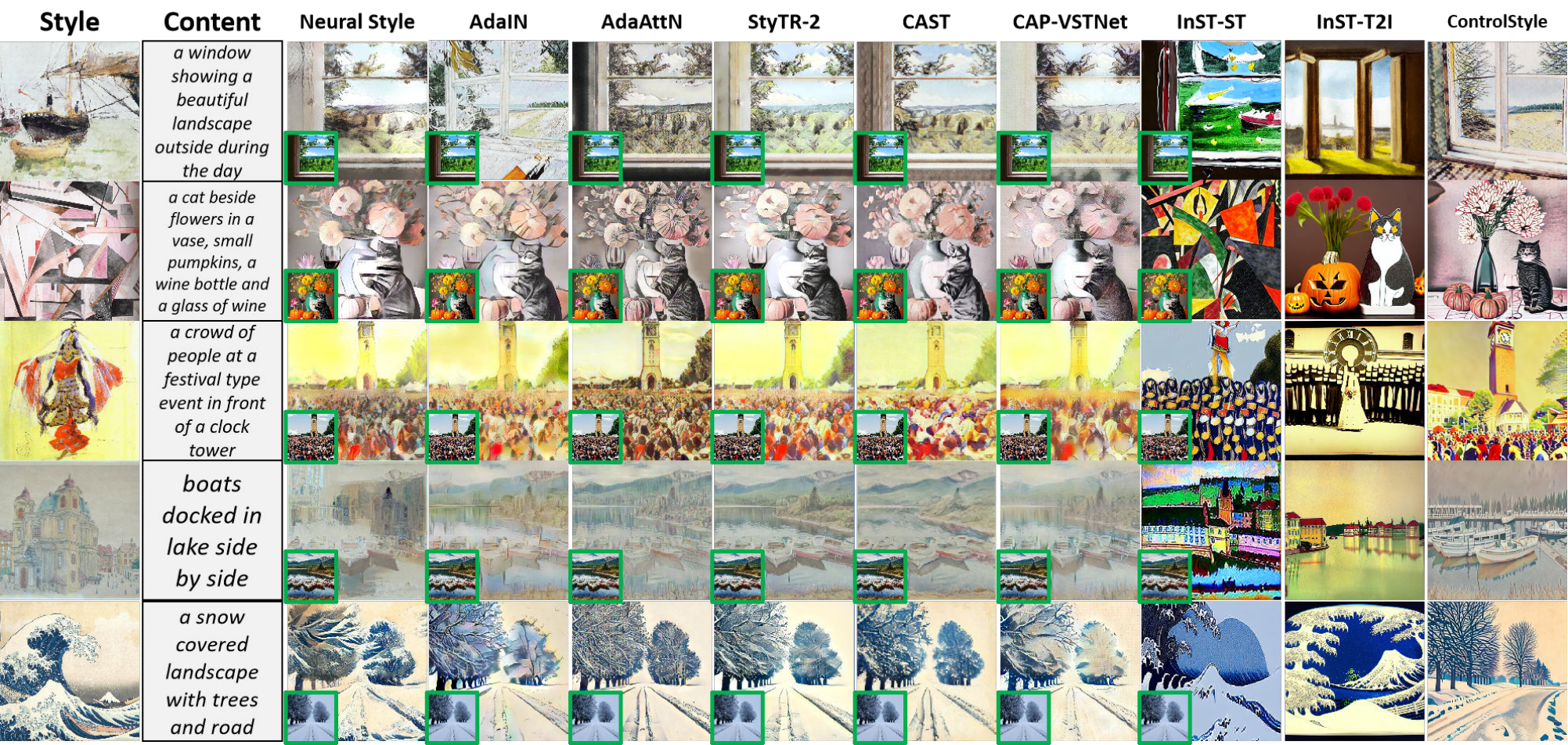}
  \vspace{-0.1in}
  \caption{Examples generated by two-stage approaches (i.e., Neural Style, AdaIN, AdaAttN, StyTR-2, CAST, CAP-VSTNet), diffusion-based InST, and our proposed ControlStyle. Please note that the two-stage methods perform the conventional style transfer on the content image generated by stable diffusion, while our ControlStyle takes an input text prompt and a style image to produce the stylized image via a unified diffusion model in a single-stage manner. Particularly, InST learns a style embedding for a target style image via text inversion and is able to perform style transfer (InST-ST) given a content image and stylized text-to-image generation (InST-T2I) given an input text prompt.}
  \label{fig:examples}
  \vspace{-0.1in}
\end{figure*}

\begin{table}[t]
\centering
\caption[]{Quantitative evaluation in text-driven stylized image generation. HPS, LAION-Aes and Human denotes the Human Preference Score, LAION Aesthetics score and user study, respectively.}
\vspace{-0.1in}
\label{tab:metric}
\begin{tabular}{llll}
\hline
Method 		& HPS  	& LAION-Aes &  Human (\%)\\ \hline
Neural Style \cite{neural_style}	&  	17.22		&  	3.91	   &    	15	\\
AdaIN \cite{adain}		&  	18.79	&  	5.52	   &  	28	\\
ControlStyle 	&  	19.09	&  	6.09	   & 	 57	\\
 \hline
\end{tabular}
\vspace{-0.1in}
\end{table}

\subsection{Performance and Comparison}
In this part, we compare our ControlStyle with two-stage (generate-then-transfer) approaches (i.e., Neural Style \cite{neural_style}, AdaIN \cite{adain}, AdaAttn \cite{adaattn}, StyTR-2 \cite{stytr2}, CAST \cite{cast} and CAP-VSTNet \cite{cap}) and diffusion-based method InST \cite{inst} (i.e., InST-ST and InST-T2I correspond to two different modes of style transfer and stylized text-to-image generation, respectively). We evaluate all the methods with captions from the validation set of MS-COCO and style images from the test set of WikiArt.

\textbf{Quantitative Comparisons.} Since there are no ground-truth stylized images available for this task, we employ two aesthetic evaluation metrics, i.e., Human Preference Score (HPS) \cite{hps} and LAION-Aesthetic score (LAION-Aes) \footnote{https://github.com/LAION-AI/aesthetic-predictor}, for quantitative comparisons among Neural Style, AdaIN and ControlStyle. HPS is pre-trained with a dataset of human choices on generated images collected from the Stable Foundation Discord channel, which can measure the alignment between images generated by text-to-image models and human aesthetic preferences. LAION-Aes is pre-trained on LAION-Aesthetics dataset to assess the aesthetic scores of the generated images in a single modality. We randomly sample 200 captions from MS-COCO and 100 style images from WikiArt for validation, leading to 20K stylized images. The random seeds for all the three methods are identical. The first two columns in Table \ref{tab:metric} show the performances of the three models. Overall, the results across these two metrics consistently indicate that our ControlStyle surpasses two-stage methods (Neural Style and AdaIN). Specifically, our ControlStyle achieves the relative improvement over Neural Style and AdaIN in LAION-Aes by 55.8\% and 10.3\%, respectively, which demonstrates the benefit of unifying the text-to-image model and style transfer network into a unified diffusion model. Since the structural fidelity is more determined by the pre-trained text-to-image diffusion model (stable diffusion), ControlStyle and AdaIN yield comparable HPS scores. Nonetheless, ControlStyle model continues to outperform the other two methods by leveraging strong image priors from the decoder in auto-encoder of stable diffusion via the proposed diffusion regularizations for text-driven stylized image generation.

\textbf{User Study.} Additionally, we conduct user study to examine whether the stylized images generated by the three methods conform to human preferences. Ten evaluators of diverse education backgrounds are invited to participate in this study, which involves 5 males and 5 females from computer science (2), art design (4), social science (2), and business (2), respectively. Evaluators are shown to the generated images by the three approaches, the corresponding text prompts and target style images, and they are asked: Which one exhibits the best visual quality and aligns well with the input text prompt and target style. The percentage of results ranking the first in the comparisons, as assessed by the evaluators, is reported. Results are listed in the last column (\emph{Human}) in Table \ref{tab:metric}, and ControlStyle outperforms the other methods by a large margin.

\textbf{Qualitative Comparisons.} In this part, we qualitatively evaluate the performances of different methods by showing some examples in Figure. \ref{fig:examples}. In general, our ControlStyle generates results more visually appealing compared to the results of the other methods. It can be observed that the spatial structures of stylized images rendered by two-stage approaches (i.e., Neural Style, AdaIN, AdaAttN, StyTR-2, CAST and CAP-VSTNet) are, to some extent, destroyed after style transfer, while our ControlStyle preserves better structures fidelity in the stylized images. The underlying principle behind this is that our ControlStyle unifies text-to-image generation and style transfer within a single diffusion model, mitigating the structure distortions that may arise in style transfer and smoothly fusing the style into the stylized images by progressive sampling. For example, our ControlStyle effectively preserves the body shape of the little cat in \emph{Row 2}, while both Neural Style and AdaIN tend to disrupt the spatial structure by directly matching features between the generated and input style images. Though the sample produced by Neural Style in \emph{Row 5} exhibits somewhat alignment with the input style image in texture, the crucial structural details for identifying trees and road are lost. In contrast, our ControlStyle generates an image that is aesthetically pleasing in both structure and style. Particularly, InST fails to generate satisfactory results and we speculate that this may be the result of using text inversion for a target style image and stochastic inversion for style transfer. Such design in InST is somewhat vulnerable to overfitting to the spatial structure in the target style image, and thus poor samples are generated when the semantics of content image and style image are completely different.

\subsection{Analysis and Discussion}
\begin{figure}[t]
  \centering
  \includegraphics[width=\linewidth]{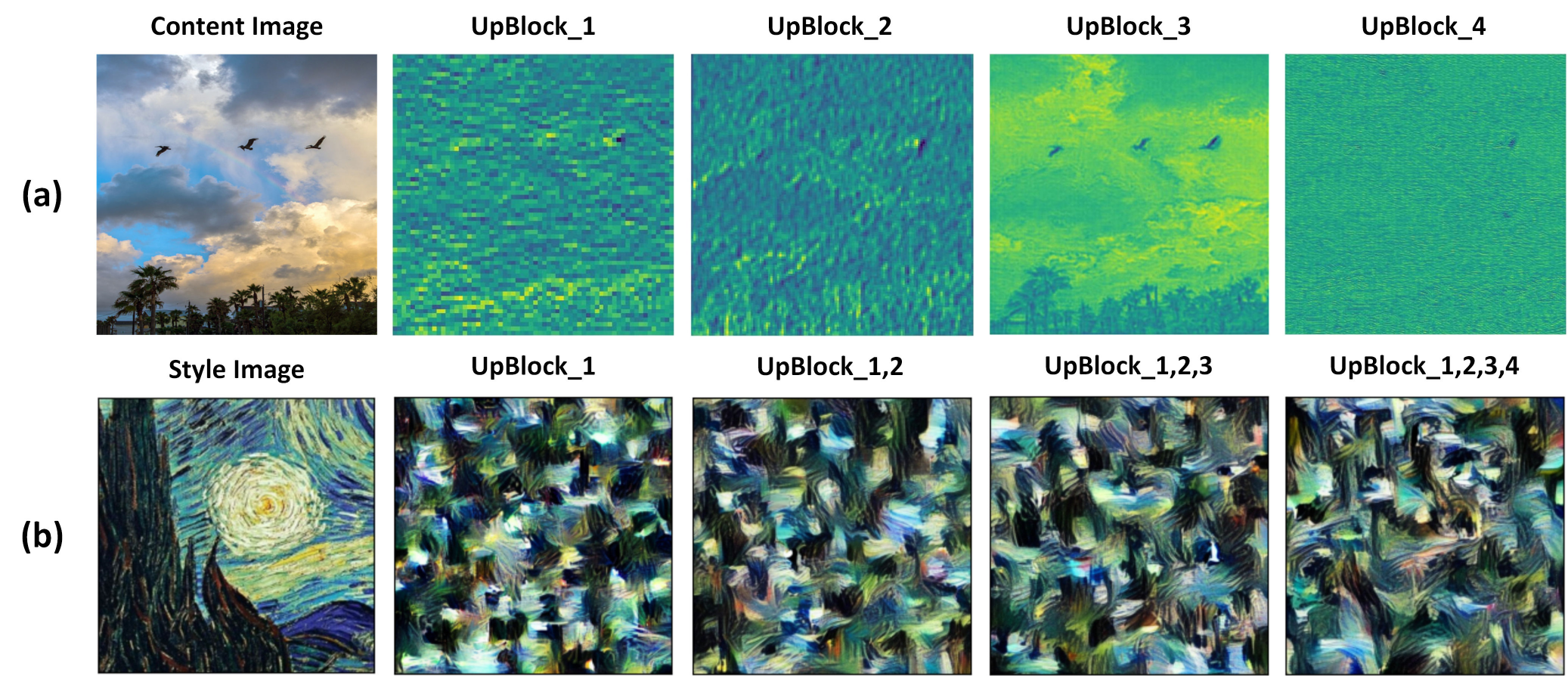}
   \vspace{-0.1in}
  \caption{(a) We try to identify the most significant features from the upsample blocks (i.e., \emph{UpBlock}) in the auto-encoder that would help ControlStyle preserve the spatial structure in the presence of diffusion style regularization through feature visualization. (b) Similar to \cite{neural_style}, we apply optimization to find out the latent code that minimizes $\mathcal{L}_{ldm}^{style}$ for several upsample blocks in the auto-encoder, and convert it back to pixel space for visualization.}
  \label{fig:feature}
  \vspace{-0.2in}
\end{figure}

\textbf{Feature Selection for Diffusion Regularizations.} To facilitate the training of our ControlStyle under unpaired setting, we devise two diffusion regularizations to align the stylized image with the content image and the style image in structure and style, respectively. In these diffusion regularizations, features from the upsample blocks in the auto-encoder of stable diffusion are utilized to measure the discrepancy between the generated image and the content/style image during training. In our experiments, indiscriminately applying the proposed diffusion regularizations to all the upsample blocks leads to degraded performances. Hence, we conduct an analysis to identify the most significant upsample blocks in terms of their features. For the diffusion content regularization, we visualize the feature maps of different upsample blocks in Figure \ref{fig:feature} (a), which shows that \emph{UpBlock\_3} effectively learns the most structure information. For the diffusion style regularization, we minimize $\mathcal{L}_{ldm}^{style}$ computed over different sets of upsample blocks to transform a randomly initialized latent code into the target style image. It can be observed that involving more blocks leads to better reconstruction. In our experiments, we remove the \emph{UpBlock\_4} in diffusion style regularization to avoid overlearning the structure of the style image, which can alleviate some artifacts and lead to smoother images.

\begin{figure}[t]
  \centering
  \includegraphics[width=0.85\linewidth]{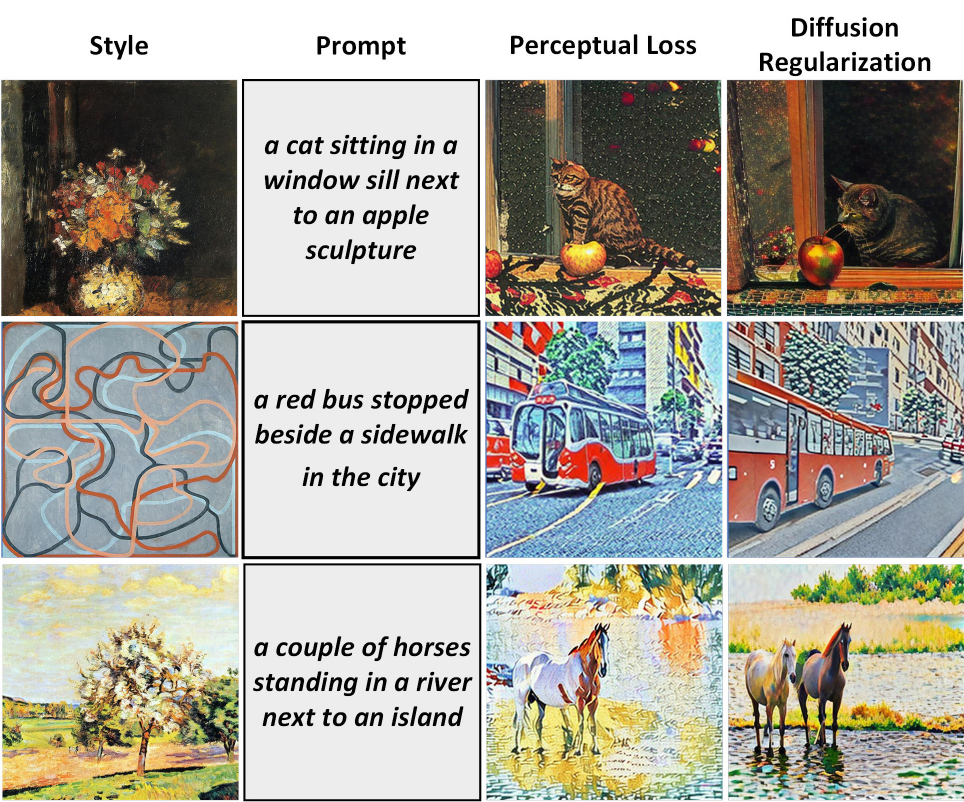}
  \vspace{-0.1in}
  \caption{Example results obtained by training the modulation network with Perceptual Loss \cite{perceptual} and our proposed Diffusion Regularizations, respectively. It can be easily observed that images generated by (ControlStyle +) Perceptual Loss suffer from more artifacts than those produced by our (ControlStyle +) Diffusion Regularizations.}
  \label{fig:loss}
  \vspace{-0.1in}
\end{figure}

\begin{figure}[t]
  \centering
  \includegraphics[width=0.85\linewidth]{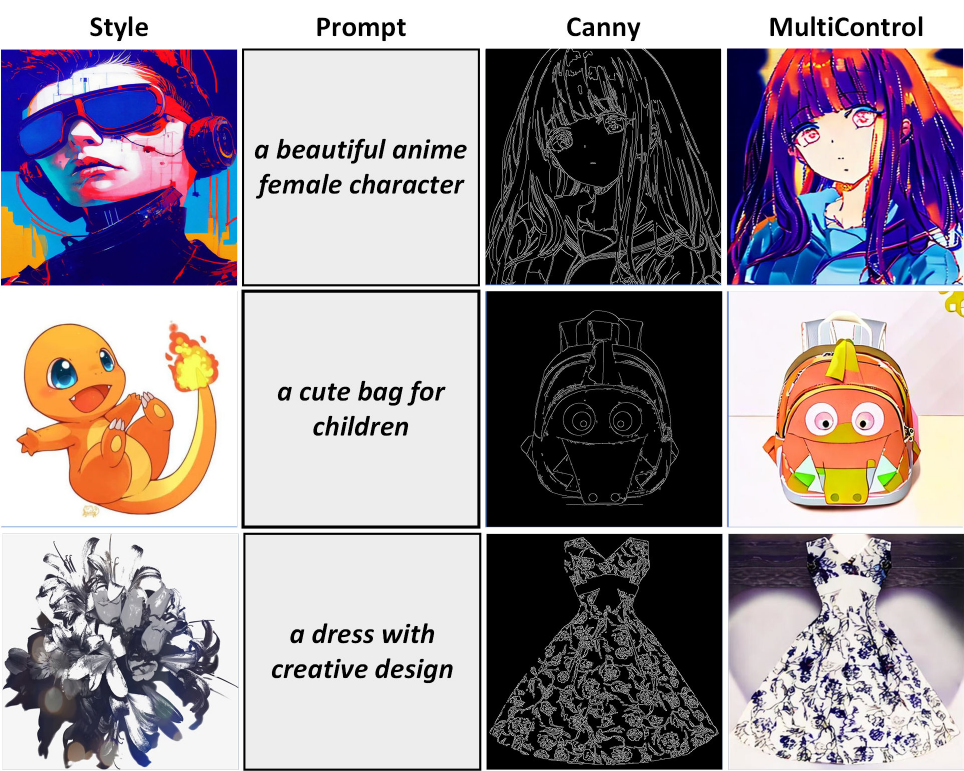}
  \vspace{-0.1in}
  \caption{Examples generated by combining our ControlStyle and a pre-trained ControlNet \cite{controlnet} with Canny edge as an additional condition. Please note that these results are achieved without retraining our ControlStyle.}
  \label{fig:multicontrol}
  \vspace{-0.1in}
\end{figure}

\begin{figure*}[t]
  \centering
  \includegraphics[width=0.85\linewidth]{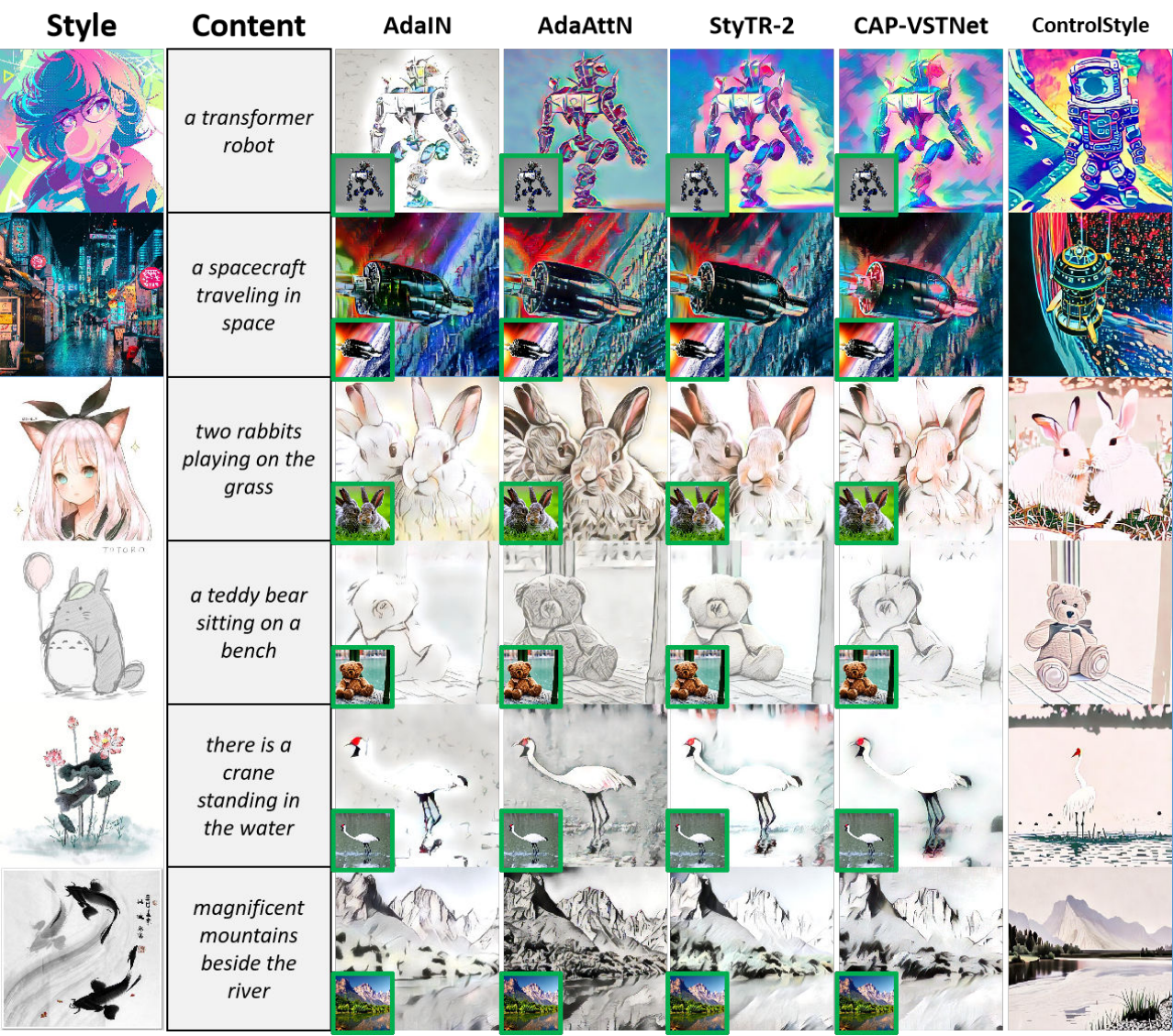}
  \vspace{-0.1in}
  \caption{Examples generated of ControlStyle and AdaIN in three styles unseen in the training data (i.e., WikiArt): cyberpunk (\emph{Row 1-2}), anime (\emph{Row 3-4}), and Chinese ink style (\emph{Row 5-6}).}
  \label{fig:ood}
   \vspace{-0.1in}
\end{figure*}

\textbf{Perceptual Loss vs Diffusion Regularizations.} Both perceptual loss \cite{perceptual} and our proposed diffusion regularizations can be employed to encourage ControlStyle to align with the style of inputs. In this part, we compare these two different training strategies and show some examples generated by these two training strategies in Figure \ref{fig:loss}. Overall, the images generated by ControlStyle trained with perceptual loss suffer from more artifacts than those generated by the model optimized with our diffusion regularizations. These results highlight the advantage of leveraging image priors from an auto-encoder pre-trained on vast amounts of data in stable diffusion. For the example of ``a red bus stopped beside a side walk in the city'' in \emph{Row 2}, more grid artifacts are observed in the image generated by \emph{(ControlStyle +) Perceptual Loss} compared with the one generated by \emph{(ControlStyle +) Diffusion Regularizations}.

\textbf{Generalizability.} Whether the trained model can be applied to styles unseen in the training dataset is a crucial factor in text-driven stylized image generation. To evaluate this capability, we compared our ControlStyle and several two-stage methods (i.e., AdaIN, AdaAttN, StyTR-2, and CAP-VSTNet) on three different unseen styles: cyberpunk, anime, and Chinese ink style. Examples are illustrated in Figure \ref{fig:ood}. AdaIN can robustly generate stylized images somewhat similar to the input style image in style. However, severe artifacts and structure distortions are introduced to these images. Instead, our ControlStyle is able to produce more impressive results than the other approaches, which demonstrates the strong generalizability of our ControlStyle. Particularly, for the style ``cyberpunk'' in \emph{Row 1-2}, ControlStyle better aligns the painting colors with the input style image and preserves better spatial structures.

\subsection{Multiple Controls}
Similar to ControlNet \cite{controlnet} that steers a pre-trained text-to-image diffusion model, our ControlStyle can also be easily extended with multiple controls in a training-free manner, leading to stronger editability for content creation. Here, we combine our ControlStyle and the pre-trained ControlNet with Canny edge as an additional condition to explore the unleashed potentials of ControlStyle in more interesting applications, such as costume or anime character design. Technically, the features from the modulation network in each diffusion model are weighted based on the corresponding control weights and aggregated. Then, the fused features are injected to the decoder of U-Net in the pre-trained stable diffusion. The control weights for our ControlStyle and ControlNet are set as 0.8 and 1.0, respectively. Some interesting examples are shown in Figure \ref{fig:multicontrol}. It is encouraging that even though our ControlStyle is not retrained along with ControlNet, promising results are attained. Particularly, both the input style image and the generated image depict red highlights on the hair in \emph{Row 1}.

\section{Conclusion}
In this paper, we develop a new task of text-driven stylized image generation, which aims to generate stylized images that are both semantically aligned with an input text prompt and consistent with a style image in style. This new task eliminates the need for a content image in stylized image generation and further enhances the editability of diffusion models for content creation. To resolve this task, we propose a new diffusion model, namely ControlStyle, to stylize a pre-trained text-to-image diffusion model (e.g., stable diffusion) with a trainable modulation network. To facilitate the training of our ControlStyle under unpaired setting, two novel diffusion regularizations are devised to enforce the target styles and prevent severe structure distortions, respectively. Extensive experiments are conducted to demonstrate the superiority of our ControlStyle in text-driven stylized image generation compared with a simple combination of a pre-trained text-to-image model and a style transfer network/algorithm. Moreover, we upgrade our ControlStyle by incorporating more controls in a training-free manner, which shows its potentials in more practical applications.


\begin{thebibliography}{47}
\balance

\ifx \showCODEN    \undefined \def \showCODEN     #1{\unskip}     \fi
\ifx \showDOI      \undefined \def \showDOI       #1{#1}\fi
\ifx \showISBNx    \undefined \def \showISBNx     #1{\unskip}     \fi
\ifx \showISBNxiii \undefined \def \showISBNxiii  #1{\unskip}     \fi
\ifx \showISSN     \undefined \def \showISSN      #1{\unskip}     \fi
\ifx \showLCCN     \undefined \def \showLCCN      #1{\unskip}     \fi
\ifx \shownote     \undefined \def \shownote      #1{#1}          \fi
\ifx \showarticletitle \undefined \def \showarticletitle #1{#1}   \fi
\ifx \showURL      \undefined \def \showURL       {\relax}        \fi
\providecommand\bibfield[2]{#2}
\providecommand\bibinfo[2]{#2}
\providecommand\natexlab[1]{#1}
\providecommand\showeprint[2][]{arXiv:#2}

\bibitem[Bai et~al\mbox{.}(2023)]%
        {tistyler}
\bibfield{author}{\bibinfo{person}{Yunpeng Bai}, \bibinfo{person}{Jiayue Liu},
  \bibinfo{person}{Chao Dong}, {and} \bibinfo{person}{Chun Yuan}.}
  \bibinfo{year}{2023}\natexlab{}.
\newblock \showarticletitle{ITstyler: Image-optimized Text-based Style
  Transfer}.
\newblock \bibinfo{journal}{\emph{CoRR}}  \bibinfo{volume}{abs/2301.10916}
  (\bibinfo{year}{2023}).
\newblock
\urldef\tempurl%
\url{https://doi.org/10.48550/arXiv.2301.10916}
\showURL{%
\tempurl}


\bibitem[Balaji et~al\mbox{.}(2022)]%
        {eDiff}
\bibfield{author}{\bibinfo{person}{Yogesh Balaji}, \bibinfo{person}{Seungjun
  Nah}, \bibinfo{person}{Xun Huang}, \bibinfo{person}{Arash Vahdat},
  \bibinfo{person}{Jiaming Song}, \bibinfo{person}{Karsten Kreis},
  \bibinfo{person}{Miika Aittala}, \bibinfo{person}{Timo Aila},
  \bibinfo{person}{Samuli Laine}, \bibinfo{person}{Bryan Catanzaro},
  \bibinfo{person}{Tero Karras}, {and} \bibinfo{person}{Ming{-}Yu Liu}.}
  \bibinfo{year}{2022}\natexlab{}.
\newblock \showarticletitle{eDiff-I: Text-to-Image Diffusion Models with an
  Ensemble of Expert Denoisers}.
\newblock \bibinfo{journal}{\emph{CoRR}}  \bibinfo{volume}{abs/2211.01324}
  (\bibinfo{year}{2022}).
\newblock
\urldef\tempurl%
\url{https://doi.org/10.48550/arXiv.2211.01324}
\showDOI{\tempurl}
\showeprint[arXiv]{2211.01324}


\bibitem[Brooks et~al\mbox{.}(2022)]%
        {instructpix2pix}
\bibfield{author}{\bibinfo{person}{Tim Brooks}, \bibinfo{person}{Aleksander
  Holynski}, {and} \bibinfo{person}{Alexei~A Efros}.}
  \bibinfo{year}{2022}\natexlab{}.
\newblock \showarticletitle{Instructpix2pix: Learning to follow image editing
  instructions}.
\newblock \bibinfo{journal}{\emph{arXiv preprint arXiv:2211.09800}}
  (\bibinfo{year}{2022}).
\newblock


\bibitem[Chen et~al\mbox{.}(2019)]%
        {chen2019mocycle}
\bibfield{author}{\bibinfo{person}{Yang Chen}, \bibinfo{person}{Yingwei Pan},
  \bibinfo{person}{Ting Yao}, \bibinfo{person}{Xinmei Tian}, {and}
  \bibinfo{person}{Tao Mei}.} \bibinfo{year}{2019}\natexlab{}.
\newblock \showarticletitle{Mocycle-gan: Unpaired video-to-video translation}.
  In \bibinfo{booktitle}{\emph{ACM MM}}.
\newblock


\bibitem[Deng et~al\mbox{.}(2022)]%
        {stytr2}
\bibfield{author}{\bibinfo{person}{Yingying Deng}, \bibinfo{person}{Fan Tang},
  \bibinfo{person}{Weiming Dong}, \bibinfo{person}{Chongyang Ma},
  \bibinfo{person}{Xingjia Pan}, \bibinfo{person}{Lei Wang}, {and}
  \bibinfo{person}{Changsheng Xu}.} \bibinfo{year}{2022}\natexlab{}.
\newblock \showarticletitle{Stytr2: Image style transfer with transformers}. In
  \bibinfo{booktitle}{\emph{CVPR}}.
\newblock


\bibitem[Dhariwal and Nichol(2021)]%
        {classifierguidance}
\bibfield{author}{\bibinfo{person}{Prafulla Dhariwal} {and}
  \bibinfo{person}{Alexander~Quinn Nichol}.} \bibinfo{year}{2021}\natexlab{}.
\newblock \showarticletitle{Diffusion Models Beat GANs on Image Synthesis}. In
  \bibinfo{booktitle}{\emph{NeurIPS}}.
\newblock


\bibitem[Fu et~al\mbox{.}(2022)]%
        {fu2022language}
\bibfield{author}{\bibinfo{person}{Tsu-Jui Fu}, \bibinfo{person}{Xin~Eric
  Wang}, {and} \bibinfo{person}{William~Yang Wang}.}
  \bibinfo{year}{2022}\natexlab{}.
\newblock \showarticletitle{Language-driven artistic style transfer}. In
  \bibinfo{booktitle}{\emph{ECCV}}.
\newblock


\bibitem[Gatys et~al\mbox{.}(2016)]%
        {neural_style}
\bibfield{author}{\bibinfo{person}{Leon~A Gatys}, \bibinfo{person}{Alexander~S
  Ecker}, {and} \bibinfo{person}{Matthias Bethge}.}
  \bibinfo{year}{2016}\natexlab{}.
\newblock \showarticletitle{Image style transfer using convolutional neural
  networks}. In \bibinfo{booktitle}{\emph{Proceedings of the IEEE conference on
  computer vision and pattern recognition}}. \bibinfo{pages}{2414--2423}.
\newblock


\bibitem[Gatys et~al\mbox{.}(2017)]%
        {gaty_control}
\bibfield{author}{\bibinfo{person}{Leon~A Gatys}, \bibinfo{person}{Alexander~S
  Ecker}, \bibinfo{person}{Matthias Bethge}, \bibinfo{person}{Aaron Hertzmann},
  {and} \bibinfo{person}{Eli Shechtman}.} \bibinfo{year}{2017}\natexlab{}.
\newblock \showarticletitle{Controlling perceptual factors in neural style
  transfer}. In \bibinfo{booktitle}{\emph{CVPR}}.
\newblock


\bibitem[Goodfellow et~al\mbox{.}(2020)]%
        {gan}
\bibfield{author}{\bibinfo{person}{Ian Goodfellow}, \bibinfo{person}{Jean
  Pouget-Abadie}, \bibinfo{person}{Mehdi Mirza}, \bibinfo{person}{Bing Xu},
  \bibinfo{person}{David Warde-Farley}, \bibinfo{person}{Sherjil Ozair},
  \bibinfo{person}{Aaron Courville}, {and} \bibinfo{person}{Yoshua Bengio}.}
  \bibinfo{year}{2020}\natexlab{}.
\newblock \showarticletitle{Generative adversarial networks}.
\newblock \bibinfo{journal}{\emph{Commun. ACM}} (\bibinfo{year}{2020}).
\newblock


\bibitem[Hertz et~al\mbox{.}(2023)]%
        {p2p}
\bibfield{author}{\bibinfo{person}{Amir Hertz}, \bibinfo{person}{Ron Mokady},
  \bibinfo{person}{Jay Tenenbaum}, \bibinfo{person}{Kfir Aberman},
  \bibinfo{person}{Yael Pritch}, {and} \bibinfo{person}{Daniel Cohen-Or}.}
  \bibinfo{year}{2023}\natexlab{}.
\newblock \showarticletitle{Prompt-to-prompt image editing with cross attention
  control}. In \bibinfo{booktitle}{\emph{ICLR}}.
\newblock


\bibitem[Ho et~al\mbox{.}(2022a)]%
        {ho2022imagenvideo}
\bibfield{author}{\bibinfo{person}{Jonathan Ho}, \bibinfo{person}{William
  Chan}, \bibinfo{person}{Chitwan Saharia}, \bibinfo{person}{Jay Whang},
  \bibinfo{person}{Ruiqi Gao}, \bibinfo{person}{Alexey Gritsenko},
  \bibinfo{person}{Diederik~P Kingma}, \bibinfo{person}{Ben Poole},
  \bibinfo{person}{Mohammad Norouzi}, \bibinfo{person}{David~J Fleet},
  {et~al\mbox{.}}} \bibinfo{year}{2022}\natexlab{a}.
\newblock \showarticletitle{Imagen video: High definition video generation with
  diffusion models}.
\newblock \bibinfo{journal}{\emph{arXiv preprint arXiv:2210.02303}}
  (\bibinfo{year}{2022}).
\newblock


\bibitem[Ho et~al\mbox{.}(2020)]%
        {ddpm:ho20}
\bibfield{author}{\bibinfo{person}{Jonathan Ho}, \bibinfo{person}{Ajay Jain},
  {and} \bibinfo{person}{Pieter Abbeel}.} \bibinfo{year}{2020}\natexlab{}.
\newblock \showarticletitle{Denoising Diffusion Probabilistic Models}. In
  \bibinfo{booktitle}{\emph{Advances in Neural Information Processing Systems
  33: Annual Conference on Neural Information Processing Systems 2020, NeurIPS
  2020, December 6-12, 2020, virtual}}, \bibfield{editor}{\bibinfo{person}{Hugo
  Larochelle}, \bibinfo{person}{Marc'Aurelio Ranzato}, \bibinfo{person}{Raia
  Hadsell}, \bibinfo{person}{Maria{-}Florina Balcan}, {and}
  \bibinfo{person}{Hsuan{-}Tien Lin}} (Eds.).
\newblock
\urldef\tempurl%
\url{https://proceedings.neurips.cc/paper/2020/hash/4c5bcfec8584af0d967f1ab10179ca4b-Abstract.html}
\showURL{%
\tempurl}


\bibitem[Ho and Salimans(2022)]%
        {ho2022classifier}
\bibfield{author}{\bibinfo{person}{Jonathan Ho} {and} \bibinfo{person}{Tim
  Salimans}.} \bibinfo{year}{2022}\natexlab{}.
\newblock \showarticletitle{Classifier-free diffusion guidance}. In
  \bibinfo{booktitle}{\emph{NeurIPS Workshop}}.
\newblock


\bibitem[Ho et~al\mbox{.}(2022b)]%
        {ho2022video}
\bibfield{author}{\bibinfo{person}{Jonathan Ho}, \bibinfo{person}{Tim
  Salimans}, \bibinfo{person}{Alexey Gritsenko}, \bibinfo{person}{William
  Chan}, \bibinfo{person}{Mohammad Norouzi}, {and} \bibinfo{person}{David~J
  Fleet}.} \bibinfo{year}{2022}\natexlab{b}.
\newblock \showarticletitle{Video diffusion models}.
\newblock \bibinfo{journal}{\emph{arXiv preprint arXiv:2204.03458}}
  (\bibinfo{year}{2022}).
\newblock


\bibitem[Huang and Belongie(2017)]%
        {adain}
\bibfield{author}{\bibinfo{person}{Xun Huang} {and} \bibinfo{person}{Serge
  Belongie}.} \bibinfo{year}{2017}\natexlab{}.
\newblock \showarticletitle{Arbitrary style transfer in real-time with adaptive
  instance normalization}. In \bibinfo{booktitle}{\emph{ICCV}}.
\newblock


\bibitem[Isola et~al\mbox{.}(2017)]%
        {pix2pix}
\bibfield{author}{\bibinfo{person}{Phillip Isola}, \bibinfo{person}{Jun-Yan
  Zhu}, \bibinfo{person}{Tinghui Zhou}, {and} \bibinfo{person}{Alexei~A
  Efros}.} \bibinfo{year}{2017}\natexlab{}.
\newblock \showarticletitle{Image-to-image translation with conditional
  adversarial networks}. In \bibinfo{booktitle}{\emph{CVPR}}.
\newblock


\bibitem[Jing et~al\mbox{.}(2020)]%
        {dain}
\bibfield{author}{\bibinfo{person}{Yongcheng Jing}, \bibinfo{person}{Xiao Liu},
  \bibinfo{person}{Yukang Ding}, \bibinfo{person}{Xinchao Wang},
  \bibinfo{person}{Errui Ding}, \bibinfo{person}{Mingli Song}, {and}
  \bibinfo{person}{Shilei Wen}.} \bibinfo{year}{2020}\natexlab{}.
\newblock \showarticletitle{Dynamic instance normalization for arbitrary style
  transfer}. In \bibinfo{booktitle}{\emph{AAAI}}.
\newblock


\bibitem[Johnson et~al\mbox{.}(2016)]%
        {perceptual}
\bibfield{author}{\bibinfo{person}{Justin Johnson}, \bibinfo{person}{Alexandre
  Alahi}, {and} \bibinfo{person}{Li Fei-Fei}.} \bibinfo{year}{2016}\natexlab{}.
\newblock \showarticletitle{Perceptual losses for real-time style transfer and
  super-resolution}. In \bibinfo{booktitle}{\emph{ECCV}}.
\newblock


\bibitem[Karayev et~al\mbox{.}(2013)]%
        {wikiart}
\bibfield{author}{\bibinfo{person}{Sergey Karayev}, \bibinfo{person}{Matthew
  Trentacoste}, \bibinfo{person}{Helen Han}, \bibinfo{person}{Aseem Agarwala},
  \bibinfo{person}{Trevor Darrell}, \bibinfo{person}{Aaron Hertzmann}, {and}
  \bibinfo{person}{Holger Winnemoeller}.} \bibinfo{year}{2013}\natexlab{}.
\newblock \showarticletitle{Recognizing image style}.
\newblock \bibinfo{journal}{\emph{arXiv preprint arXiv:1311.3715}}
  (\bibinfo{year}{2013}).
\newblock


\bibitem[Kim and Ye(2021)]%
        {kim2021diffusionclip}
\bibfield{author}{\bibinfo{person}{Gwanghyun Kim} {and}
  \bibinfo{person}{Jong~Chul Ye}.} \bibinfo{year}{2021}\natexlab{}.
\newblock \showarticletitle{Diffusionclip: Text-guided image manipulation using
  diffusion models}.
\newblock  (\bibinfo{year}{2021}).
\newblock


\bibitem[Kingma and Ba(2014)]%
        {adam}
\bibfield{author}{\bibinfo{person}{Diederik~P Kingma} {and}
  \bibinfo{person}{Jimmy Ba}.} \bibinfo{year}{2014}\natexlab{}.
\newblock \showarticletitle{Adam: A method for stochastic optimization}.
\newblock \bibinfo{journal}{\emph{arXiv preprint arXiv:1412.6980}}
  (\bibinfo{year}{2014}).
\newblock


\bibitem[Kwon and Ye(2021)]%
        {kwon2021clipstyler}
\bibfield{author}{\bibinfo{person}{Gihyun Kwon} {and}
  \bibinfo{person}{Jong~Chul Ye}.} \bibinfo{year}{2021}\natexlab{}.
\newblock \showarticletitle{Clipstyler: Image style transfer with a single text
  condition}.
\newblock \bibinfo{journal}{\emph{arXiv preprint arXiv:2112.00374}}
  (\bibinfo{year}{2021}).
\newblock


\bibitem[Li et~al\mbox{.}(2022)]%
        {li2022contextual}
\bibfield{author}{\bibinfo{person}{Yehao Li}, \bibinfo{person}{Ting Yao},
  \bibinfo{person}{Yingwei Pan}, {and} \bibinfo{person}{Tao Mei}.}
  \bibinfo{year}{2022}\natexlab{}.
\newblock \showarticletitle{Contextual transformer networks for visual
  recognition}.
\newblock \bibinfo{journal}{\emph{IEEE TPAMI}} (\bibinfo{year}{2022}).
\newblock


\bibitem[Lin et~al\mbox{.}(2014)]%
        {mscoco}
\bibfield{author}{\bibinfo{person}{Tsung-Yi Lin}, \bibinfo{person}{Michael
  Maire}, \bibinfo{person}{Serge Belongie}, \bibinfo{person}{James Hays},
  \bibinfo{person}{Pietro Perona}, \bibinfo{person}{Deva Ramanan},
  \bibinfo{person}{Piotr Doll{\'a}r}, {and} \bibinfo{person}{C~Lawrence
  Zitnick}.} \bibinfo{year}{2014}\natexlab{}.
\newblock \showarticletitle{Microsoft coco: Common objects in context}. In
  \bibinfo{booktitle}{\emph{ECCV}}.
\newblock


\bibitem[Liu et~al\mbox{.}(2021)]%
        {adaattn}
\bibfield{author}{\bibinfo{person}{Songhua Liu}, \bibinfo{person}{Tianwei Lin},
  \bibinfo{person}{Dongliang He}, \bibinfo{person}{Fu Li},
  \bibinfo{person}{Meiling Wang}, \bibinfo{person}{Xin Li},
  \bibinfo{person}{Zhengxing Sun}, \bibinfo{person}{Qian Li}, {and}
  \bibinfo{person}{Errui Ding}.} \bibinfo{year}{2021}\natexlab{}.
\newblock \showarticletitle{Adaattn: Revisit attention mechanism in arbitrary
  neural style transfer}. In \bibinfo{booktitle}{\emph{ICCV}}.
\newblock


\bibitem[Mirza and Osindero(2014)]%
        {cgan}
\bibfield{author}{\bibinfo{person}{Mehdi Mirza} {and} \bibinfo{person}{Simon
  Osindero}.} \bibinfo{year}{2014}\natexlab{}.
\newblock \showarticletitle{Conditional generative adversarial nets}.
\newblock \bibinfo{journal}{\emph{arXiv preprint arXiv:1411.1784}}
  (\bibinfo{year}{2014}).
\newblock


\bibitem[Nichol et~al\mbox{.}(2021)]%
        {glide}
\bibfield{author}{\bibinfo{person}{Alex Nichol}, \bibinfo{person}{Prafulla
  Dhariwal}, \bibinfo{person}{Aditya Ramesh}, \bibinfo{person}{Pranav Shyam},
  \bibinfo{person}{Pamela Mishkin}, \bibinfo{person}{Bob McGrew},
  \bibinfo{person}{Ilya Sutskever}, {and} \bibinfo{person}{Mark Chen}.}
  \bibinfo{year}{2021}\natexlab{}.
\newblock \showarticletitle{Glide: Towards photorealistic image generation and
  editing with text-guided diffusion models}.
\newblock \bibinfo{journal}{\emph{arXiv preprint arXiv:2112.10741}}
  (\bibinfo{year}{2021}).
\newblock


\bibitem[Pan et~al\mbox{.}(2017)]%
        {pan2017create}
\bibfield{author}{\bibinfo{person}{Yingwei Pan}, \bibinfo{person}{Zhaofan Qiu},
  \bibinfo{person}{Ting Yao}, \bibinfo{person}{Houqiang Li}, {and}
  \bibinfo{person}{Tao Mei}.} \bibinfo{year}{2017}\natexlab{}.
\newblock \showarticletitle{To create what you tell: Generating videos from
  captions}. In \bibinfo{booktitle}{\emph{ACM MM}}.
\newblock


\bibitem[Park et~al\mbox{.}(2020)]%
        {cut}
\bibfield{author}{\bibinfo{person}{Taesung Park}, \bibinfo{person}{Alexei~A
  Efros}, \bibinfo{person}{Richard Zhang}, {and} \bibinfo{person}{Jun-Yan
  Zhu}.} \bibinfo{year}{2020}\natexlab{}.
\newblock \showarticletitle{Contrastive learning for unpaired image-to-image
  translation}. In \bibinfo{booktitle}{\emph{ECCV}}.
\newblock


\bibitem[Poole et~al\mbox{.}(2022)]%
        {dreamfusion}
\bibfield{author}{\bibinfo{person}{Ben Poole}, \bibinfo{person}{Ajay Jain},
  \bibinfo{person}{Jonathan~T Barron}, {and} \bibinfo{person}{Ben Mildenhall}.}
  \bibinfo{year}{2022}\natexlab{}.
\newblock \showarticletitle{Dreamfusion: Text-to-3d using 2d diffusion}.
\newblock \bibinfo{journal}{\emph{arXiv preprint arXiv:2209.14988}}
  (\bibinfo{year}{2022}).
\newblock


\bibitem[Radford et~al\mbox{.}(2021)]%
        {clip}
\bibfield{author}{\bibinfo{person}{Alec Radford}, \bibinfo{person}{Jong~Wook
  Kim}, \bibinfo{person}{Chris Hallacy}, \bibinfo{person}{Aditya Ramesh},
  \bibinfo{person}{Gabriel Goh}, \bibinfo{person}{Sandhini Agarwal},
  \bibinfo{person}{Girish Sastry}, \bibinfo{person}{Amanda Askell},
  \bibinfo{person}{Pamela Mishkin}, \bibinfo{person}{Jack Clark},
  \bibinfo{person}{Gretchen Krueger}, {and} \bibinfo{person}{Ilya Sutskever}.}
  \bibinfo{year}{2021}\natexlab{}.
\newblock \showarticletitle{Learning Transferable Visual Models From Natural
  Language Supervision}. In \bibinfo{booktitle}{\emph{ICML}}.
\newblock


\bibitem[Rombach et~al\mbox{.}(2022)]%
        {ldm:cvpr22}
\bibfield{author}{\bibinfo{person}{Robin Rombach}, \bibinfo{person}{Andreas
  Blattmann}, \bibinfo{person}{Dominik Lorenz}, \bibinfo{person}{Patrick
  Esser}, {and} \bibinfo{person}{Bj{\"{o}}rn Ommer}.}
  \bibinfo{year}{2022}\natexlab{}.
\newblock \showarticletitle{High-Resolution Image Synthesis with Latent
  Diffusion Models}. In \bibinfo{booktitle}{\emph{CVPR}}.
\newblock


\bibitem[Ronneberger et~al\mbox{.}(2015)]%
        {unet}
\bibfield{author}{\bibinfo{person}{Olaf Ronneberger}, \bibinfo{person}{Philipp
  Fischer}, {and} \bibinfo{person}{Thomas Brox}.}
  \bibinfo{year}{2015}\natexlab{}.
\newblock \showarticletitle{U-net: Convolutional networks for biomedical image
  segmentation}. In \bibinfo{booktitle}{\emph{MICCAI}}.
\newblock


\bibitem[Saharia et~al\mbox{.}(2022)]%
        {imagen}
\bibfield{author}{\bibinfo{person}{Chitwan Saharia}, \bibinfo{person}{William
  Chan}, \bibinfo{person}{Saurabh Saxena}, \bibinfo{person}{Lala Li},
  \bibinfo{person}{Jay Whang}, \bibinfo{person}{Emily Denton},
  \bibinfo{person}{Seyed Kamyar~Seyed Ghasemipour},
  \bibinfo{person}{Burcu~Karagol Ayan}, \bibinfo{person}{S.~Sara Mahdavi},
  \bibinfo{person}{Rapha~Gontijo Lopes}, \bibinfo{person}{Tim Salimans},
  \bibinfo{person}{Jonathan Ho}, \bibinfo{person}{David~J. Fleet}, {and}
  \bibinfo{person}{Mohammad Norouzi}.} \bibinfo{year}{2022}\natexlab{}.
\newblock \showarticletitle{Photorealistic Text-to-Image Diffusion Models with
  Deep Language Understanding}.
\newblock \bibinfo{journal}{\emph{CoRR}}  \bibinfo{volume}{abs/2205.11487}
  (\bibinfo{year}{2022}).
\newblock
\urldef\tempurl%
\url{https://doi.org/10.48550/arXiv.2205.11487}
\showDOI{\tempurl}
\showeprint[arXiv]{2205.11487}


\bibitem[Song et~al\mbox{.}(2020)]%
        {ddim}
\bibfield{author}{\bibinfo{person}{Jiaming Song}, \bibinfo{person}{Chenlin
  Meng}, {and} \bibinfo{person}{Stefano Ermon}.}
  \bibinfo{year}{2020}\natexlab{}.
\newblock \showarticletitle{Denoising diffusion implicit models}.
\newblock \bibinfo{journal}{\emph{arXiv preprint arXiv:2010.02502}}
  (\bibinfo{year}{2020}).
\newblock


\bibitem[Tang et~al\mbox{.}(2023)]%
        {makeit3d}
\bibfield{author}{\bibinfo{person}{Junshu Tang}, \bibinfo{person}{Tengfei
  Wang}, \bibinfo{person}{Bo Zhang}, \bibinfo{person}{Ting Zhang},
  \bibinfo{person}{Ran Yi}, \bibinfo{person}{Lizhuang Ma}, {and}
  \bibinfo{person}{Dong Chen}.} \bibinfo{year}{2023}\natexlab{}.
\newblock \showarticletitle{Make-It-3D: High-Fidelity 3D Creation from A Single
  Image with Diffusion Prior}.
\newblock \bibinfo{journal}{\emph{arXiv preprint arXiv:2303.14184}}
  (\bibinfo{year}{2023}).
\newblock


\bibitem[Wang et~al\mbox{.}(2018)]%
        {wang2018pix2pixHD}
\bibfield{author}{\bibinfo{person}{Ting-Chun Wang}, \bibinfo{person}{Ming-Yu
  Liu}, \bibinfo{person}{Jun-Yan Zhu}, \bibinfo{person}{Andrew Tao},
  \bibinfo{person}{Jan Kautz}, {and} \bibinfo{person}{Bryan Catanzaro}.}
  \bibinfo{year}{2018}\natexlab{}.
\newblock \showarticletitle{High-Resolution Image Synthesis and Semantic
  Manipulation with Conditional GANs}. In \bibinfo{booktitle}{\emph{CVPR}}.
\newblock


\bibitem[Wen et~al\mbox{.}(2023)]%
        {cap}
\bibfield{author}{\bibinfo{person}{Linfeng Wen}, \bibinfo{person}{Chengying
  Gao}, {and} \bibinfo{person}{Changqing Zou}.}
  \bibinfo{year}{2023}\natexlab{}.
\newblock \showarticletitle{CAP-VSTNet: Content Affinity Preserved Versatile
  Style Transfer}. In \bibinfo{booktitle}{\emph{CVPR}}.
\newblock


\bibitem[Wu et~al\mbox{.}(2023)]%
        {hps}
\bibfield{author}{\bibinfo{person}{Xiaoshi Wu}, \bibinfo{person}{Keqiang Sun},
  \bibinfo{person}{Feng Zhu}, \bibinfo{person}{Rui Zhao}, {and}
  \bibinfo{person}{Hongsheng Li}.} \bibinfo{year}{2023}\natexlab{}.
\newblock \showarticletitle{Better Aligning Text-to-Image Models with Human
  Preference}.
\newblock \bibinfo{journal}{\emph{arXiv preprint arXiv:2303.14420}}
  (\bibinfo{year}{2023}).
\newblock


\bibitem[Yang et~al\mbox{.}(2023)]%
        {zeroshot_tistyler}
\bibfield{author}{\bibinfo{person}{Serin Yang}, \bibinfo{person}{Hyunmin
  Hwang}, {and} \bibinfo{person}{Jong~Chul Ye}.}
  \bibinfo{year}{2023}\natexlab{}.
\newblock \showarticletitle{Zero-Shot Contrastive Loss for Text-Guided
  Diffusion Image Style Transfer}.
\newblock \bibinfo{journal}{\emph{CoRR}}  \bibinfo{volume}{abs/2303.08622}
  (\bibinfo{year}{2023}).
\newblock


\bibitem[Yao et~al\mbox{.}(2023)]%
        {yao2023dual}
\bibfield{author}{\bibinfo{person}{Ting Yao}, \bibinfo{person}{Yehao Li},
  \bibinfo{person}{Yingwei Pan}, \bibinfo{person}{Yu Wang},
  \bibinfo{person}{Xiao-Ping Zhang}, {and} \bibinfo{person}{Tao Mei}.}
  \bibinfo{year}{2023}\natexlab{}.
\newblock \showarticletitle{Dual vision transformer}.
\newblock \bibinfo{journal}{\emph{IEEE TPAMI}} (\bibinfo{year}{2023}).
\newblock


\bibitem[Yao et~al\mbox{.}(2022)]%
        {yao2022wave}
\bibfield{author}{\bibinfo{person}{Ting Yao}, \bibinfo{person}{Yingwei Pan},
  \bibinfo{person}{Yehao Li}, \bibinfo{person}{Chong-Wah Ngo}, {and}
  \bibinfo{person}{Tao Mei}.} \bibinfo{year}{2022}\natexlab{}.
\newblock \showarticletitle{Wave-vit: Unifying wavelet and transformers for
  visual representation learning}. In \bibinfo{booktitle}{\emph{ECCV}}.
\newblock


\bibitem[Zhang and Agrawala(2023)]%
        {controlnet}
\bibfield{author}{\bibinfo{person}{Lvmin Zhang} {and} \bibinfo{person}{Maneesh
  Agrawala}.} \bibinfo{year}{2023}\natexlab{}.
\newblock \bibinfo{title}{Adding Conditional Control to Text-to-Image Diffusion
  Models}.
\newblock
\newblock
\showeprint[arxiv]{2302.05543}~[cs.CV]


\bibitem[Zhang et~al\mbox{.}(2023)]%
        {inst}
\bibfield{author}{\bibinfo{person}{Yuxin Zhang}, \bibinfo{person}{Nisha Huang},
  \bibinfo{person}{Fan Tang}, \bibinfo{person}{Haibin Huang},
  \bibinfo{person}{Chongyang Ma}, \bibinfo{person}{Weiming Dong}, {and}
  \bibinfo{person}{Changsheng Xu}.} \bibinfo{year}{2023}\natexlab{}.
\newblock \showarticletitle{Inversion-based style transfer with diffusion
  models}. In \bibinfo{booktitle}{\emph{CVPR}}.
\newblock


\bibitem[Zhang et~al\mbox{.}(2022)]%
        {cast}
\bibfield{author}{\bibinfo{person}{Yuxin Zhang}, \bibinfo{person}{Fan Tang},
  \bibinfo{person}{Weiming Dong}, \bibinfo{person}{Haibin Huang},
  \bibinfo{person}{Chongyang Ma}, \bibinfo{person}{Tong-Yee Lee}, {and}
  \bibinfo{person}{Changsheng Xu}.} \bibinfo{year}{2022}\natexlab{}.
\newblock \showarticletitle{Domain enhanced arbitrary image style transfer via
  contrastive learning}. In \bibinfo{booktitle}{\emph{ACM SIGGRAPH}}.
\newblock


\bibitem[Zhu et~al\mbox{.}(2017)]%
        {cyclegan}
\bibfield{author}{\bibinfo{person}{Jun-Yan Zhu}, \bibinfo{person}{Taesung
  Park}, \bibinfo{person}{Phillip Isola}, {and} \bibinfo{person}{Alexei~A
  Efros}.} \bibinfo{year}{2017}\natexlab{}.
\newblock \showarticletitle{Unpaired image-to-image translation using
  cycle-consistent adversarial networks}. In \bibinfo{booktitle}{\emph{ICCV}}.
\newblock


\end{thebibliography}

\end{document}